\def\year{2022}\relax
\documentclass[letterpaper]{article} 
\usepackage{aaai22}  
\usepackage{times}  
\usepackage{helvet}  
\usepackage{courier}  
\usepackage[hyphens]{url}  
\usepackage{graphicx} 
\usepackage{natbib}  
\usepackage{caption} 

\DeclareCaptionStyle{ruled}{labelfont=normalfont,labelsep=colon,strut=off} 
\usepackage{color}
\usepackage{overpic}
\usepackage{verbatim}
\usepackage{graphicx}
\usepackage{wrapfig}
\usepackage{amsmath}
\usepackage{booktabs}
\usepackage{diagbox}
\usepackage{multirow}
\usepackage{arydshln}
\usepackage[noend]{algpseudocode}
\usepackage{amssymb}
\usepackage[dvipsnames]{xcolor}
\usepackage{algorithmicx,algorithm}

\usepackage{multicol}

\usepackage{ntheorem}
\theoremseparator{:}
\newtheorem{discovery}{Discovery}

\usepackage{newfloat}
\usepackage{listings}
\floatstyle{ruled}
\newfloat{listing}{tb}{lst}{}
\floatname{listing}{Listing}
\title{Model Doctor: A Simple Gradient Aggregation Strategy for Diagnosing and Treating CNN Classifiers}
\author{
    Zunlei Feng, Jiacong Hu, Sai Wu, Xiaotian Yu, Jie Song, Mingli Song\thanks{Corresponding author.}\\
}
\affiliations{
    Zhejiang University\\
    \{zunleifeng, jiaconghu, wusai, yuxiaotian, sjie, brooksong\}@zju.edu.cn
}

\usepackage{bibentry}

\begin{document}

\def\mathbi#1{\textbf{\em #1}}

\maketitle

\begin{abstract}
Recently, Convolutional Neural Network (CNN) has achieved excellent performance in the classification task.
It is widely known that CNN is deemed as a `black-box', which is hard for understanding the prediction mechanism and debugging the wrong prediction.
Some model debugging and explanation works are developed for solving the above drawbacks.
However, those methods focus on explanation and diagnosing possible causes for model prediction, based on which the researchers handle the following optimization of models manually.
In this paper, we propose the first completely automatic model diagnosing and treating tool, termed as Model Doctor.
Based on two discoveries that 1) each category is only correlated with sparse and specific convolution kernels, and 2) adversarial samples are isolated while normal samples are successive in the feature space,
 a simple aggregate gradient constraint is devised for effectively diagnosing and optimizing CNN classifiers.
The aggregate gradient strategy is a versatile module for mainstream CNN classifiers.
Extensive experiments demonstrate that the proposed Model Doctor applies to all existing CNN classifiers,
 and improves the accuracy of $16$ mainstream CNN classifiers by $1\%\sim 5\%$.

\end{abstract}

\section{Introduction}

Image classification is a widely studied topic in the computer vision area.
The image classification algorithms are the essential ingredient in many applications,
such as, face recognition~\cite{2018Deep}, autonomous driving~\cite{DBLP:journals/corr/BojarskiTDFFGJM16}, text recognition~\cite{DBLP:journals/corr/abs-2005-03492,2015Text}, medical image analysis~\cite{2017A,2021Edge-competing}, product quality inspection~\cite{2018Scale} and so on.
In recent years, Convolutional Neural Network (CNN) based classification networks,
e.g.,
AlexNet~\cite{2012ImageNet},
VGG-Net~\cite{2014Very},
ResNet~\cite{2016Deep},
DenseNet~\cite{2017Densely},
SimpleNet~\cite{hasanpour2016lets},
GoogLeNet~\cite{2014Going},
MobileNet~\cite{2017MobileNets},
ShuffleNet~\cite{2017ShuffleNet},
SqueezeNet~\cite{2016SqueezeNet:},
MnasNet~\cite{tan2019mnasnet},
have achieved breakthrough results across a variety of image classification tasks.

Those CNN classifiers achieve high classification performance but often lack straightforward interpretability of the classifier predictions~\cite{2018Understanding,2018Explanation,2018A}.
In other words, the classifier acts as a blackbox and does not provide details about why it reaches a specific classification decision.

Some works are developed to explain and debug the machine learning models \cite{shouling2019survey}.
Most current works \cite{krause2016interacting,krause2017a,zhang2019manifold} focus on explaining and diagnosing the deep learning model's prediction with visual analysis techniques.
Based on those visualized diagnoses and analyses, researchers interactively optimize the deep learning models through improving data quality, choosing reliable features, and fine-tuning the model parameters.
Another kind of works \cite{bastani2017interpretability,lei2018understanding} adopted the explainable random forest to approximate the deep learning models and  analyze the deep learning model by examining the explainable model.
To sum up, all existing deep model explanation and debugging methods require humans interaction.
There is a lack of an entirely automatic model diagnosing and treating tool for efficiently and effectively optimize the deep learning models.

In this paper, we put forward a completely automatic model diagnosing and treating tool, termed as Model Doctor, which is based on two discoveries.
The first discovery is that the target category is highly correlated with sparse and specific convolutional kernels in the last few layers of the CNN classifier.
Meanwhile, there are also some incorrect responses for those corresponding convolutional kernels in the background areas.
The second discovery is that adversarial samples are isolated while normal samples are successive in the feature space.
So, we devise the gradient aggregation strategy for diagnosing the deficiency of CNN classifiers and treating them automatically.

In the diagnosing stage, the absolute value of the first-order derivatives of the predict category w.r.t. each feature map of each layer are summed into a single correlation index,
which denotes the correlation degree between the target category and the corresponding convolutional kernel.
To alleviate the disturbance of inaccurate features,
we also calculate the relevance index for the disturbed feature maps that are synthesized by adding some small noises to the original feature maps.
Then, we aggregate those correlation index values in each layer of the CNN classifier for all training samples of the same category.
The aggregated correlation index values for each convolutional kernel can be seen as the criterion for diagnosing the wrong prediction samples.

In  the treating stage,
we devise the channel-wise and space-wise constraints for treating CNN classifiers based on the above correlated convolutional kernels distribution for each category.
The channel-wise constraint is adopted to restrain the wrong correlation between the convolutional kernels and the target category.
The space-wise constraint is adopted to restrain the wrong correlation between the unrelated background features and the target category, which requires extra coarse annotations of the object area.
Experiments demonstrate that the proposed Model Doctor can effectively diagnose the possible causes for the failure prediction of the model and improve the accuracy of the CNN classifier effectively.

Our contribution is therefore the first completely automatic model diagnosing and treating tool, termed as Model Doctor.
We reveal two discoveries and analyze the relationship between the convolution kernels and categories, which can be used as the criterion for future researches on deep model diagnosing and treating.
The gradient aggregation strategy combined with the channel-wise and space-wise constraints is devised for diagnosing and treating CNN classifiers.
Extensive experiments demonstrate that the proposed methods effectively improve the accuracy of mainstream CNN classifiers by $1\% \sim 5\%$.
It's worth noting that the proposed Model Doctor is built on top of pre-trained CNN classifiers, which is applicable for all existing CNN classifiers.

\subsection{Related Work}

\subsubsection{Model Debugging and Explanation.}
Due to the complex operation mechanism and low transparency of machine learning models, there is usually a lack of reliable causes and reasoning to assist researchers in debugging the models.
Some model explanation techniques are developed as tools for model debugging and analysis.
\citet{cadamuro2016debugging} proposed a debugging method for machine learning models to identify the training items most responsible for biasing the model towards creating this error.
\citet{krause2016interacting} presented an explanatory debugging method that explains to users how the system made each of its predictions. The user then explains any necessary corrections back to the learning system.
\citet{kulesza2010explanatory} presented an explanatory debugging approach for debugging machine-learned programs.
\citet{brooks2015featureinsight} presented an interactive visual analytics tool FeatureInsight for building new dictionary features (semantically related groups of words) for text classification problems.
\citet{paiva2015an} proposed a visual data classification methodology that supports users in tasks related to categorization such as training set selection, model creation, verification, and classifier tuning.
The above methods focused on debugging traditional machine learning models. The improvement process needs human interaction.

For the deep learning models, \citet{bastani2017interpretability,lei2018understanding} adopted the explainable random forests to approximate the blackbox model and debug the blackbox model by examining the explainable model.
\citet{krause2016interacting} proposed an interactive partial dependence diagnostics to understand how features affect the prediction overall.
\citet{krause2017a} proposed the visual analytics workflow that leverages ¡°instance-level explanations¡±, measures of local feature relevance that explain single instances.
\citet{zhang2019manifold} proposed a framework that utilizes visual analysis techniques to support the interpretation, debugging and comparison of machine learning models in an interactive manner.
Those methods focused on diagnosing the possible causes for models' prediction, based on which researchers handle the following model debugging and optimization manually.
Different from the above methods, we focus on automatically treating the deep models based on the diagnosed results.

\subsubsection{Attribution Method.}
Existing attribution methods contain perturbation-based and backpropagation-based methods.
Perturbation-based methods \cite{zeiler2014visualizing,zhou2015predicting,zintgraf2017visualizing,lengerich2017visual} directly compute the attribution of an input feature by removing, masking, or altering them and running a forward pass on the new input, measuring the difference with the actual output.

For the backpropagation-based method,
\citet{baehrens2010how} first applied the first-order derivative of the predict category w.r.t. the input to explain classification decisions of the Bayesian classification task.
Furthermore,
\citet{simonyan2013deep} extended the same technique into the CNN classification network to extract class aware saliency maps.
\citet{sundararajan2017axiomatic} took the (signed) partial derivatives of the output w.r.t the input and multiplying them with the input itself (i.e., Gradient $\times$ Input),
which improves the sharpness of the attribution maps.
\citet{shrikumar2016not} computed the average gradient while the input varies along a linear path from a baseline, which is defined by the user and often chosen to be zero, to the initial input (i.e., Integrated Gradients).
\citet{shrikumar2017learning} recently proposed DeepLift for decomposing the output prediction of a neural network on a specific input by backpropagating the contributions of all neurons in the network to every feature of the input.
 \citet{bach2015on,feng2018joint} proposed an approach for propagating importance scores called Layerwise Relevance Propagation (LRP).
 \citet{kindermans2016investigating} showed that the original LRP rules were equivalent within a scaling factor to the gradient $\times$ input.

With the above techniques,
\citet{song2019deep} adopted the Saliency Map, Gradient $\times$ Input, and $\epsilon$-LRP to calculate the attribution map for estimating the transferability of deep networks.
\citet{ancona2018towards}  analyzed above Gradient $\times$ Input, $\epsilon$-LRP, Integrated Gradients and DeepLIFT (rescale) from theoretical and practical perspectives,
which shows that these four methods, despite their different formulation, are firmly related, proving conditions of equivalence or approximation between them.

\subsubsection{CAM and Grad-CAM.} Another related area is visual feature localization, which includes gradient-free technique and gradient-based technique.
For the gradient-free technique,
\citet{2016Learning} proposed the first visual feature localization technique, called Class Activation Mapping (CAM), to visualize the predicted class scores on any given image, highlighting
the discriminative object parts detected by the CNN. CAM computes a weighted sum of the feature maps of the last convolutional layer to obtain the class activation maps.
\citet{wang2020score} proposed the score-CAM,  a novel CAM variant, which uses the increase in confidence for the weight of each activation map.
\citet{desai2020ablation} introduced an enhanced visual explanation for visual sharpness called SS-CAM, which produces centralized localization of object features within an image through a smooth operation.
\citet{desai2020ablation} proposed the Ablation-CAM that uses ablation analysis to determine the importance (weights) of individual feature map units w.r.t. class, which is time-consuming.

For the gradient-based technique,
\citet{selvaraju2020grad} proposed the Grad-CAM, which utilizes a local gradient to represent the linear weight and can be applied to any average pooling-based CNN architectures without re-training.
\citet{chattopadhay2018grad} proposed the Grad-CAM++ that used a weighted combination of the positive high order derivatives of the last convolutional layer feature maps w.r.t a specific class score as weights to generate better object localization as well as explaining occurrences of multiple object instances in a single image.
\citet{omeiza2019smooth} proposed the Smooth Grad-CAM++ that calculates these maps by averaging gradients from many small perturbations of a given image and applying the resulting gradients in the generalized Grad-CAM algorithm.

\subsubsection{Deconvolutional Visualization.}
The deconvolutional network \cite{zeiler2010deconvolutional,zeiler2011adaptive} was originally proposed to learning representation in an unsupervised manner and later applied to visualization \cite{2014Visualizing}.
\citet{2014Visualizing} proposed the first deconvolution visualization approach DeConvNet to better understand what the higher layers in a given network have learned. ¡°DeConvNet¡± makes data flow from a neuron activation in the higher layers down to the image.
\citet{JT2014STRIVING} extended this work to guided backpropagation which helped understand the impact of each neuron in a deep network w.r.t. the input image.
\citet{mahendran2016salient}  extended DeConvNet to a general method for architecture reversal and visualization.






\section{Method}

\subsection{Discovery and Analysis}

\begin{discovery}
The target category is only correlated with sparse and specific convolutional kernels in the last few layers of the CNN classifier.
Meanwhile, unrelated background features will disturb the prediction of the CNN classifier.
\end{discovery}

\begin{figure}[!t]
\centering
\includegraphics[scale =0.63]{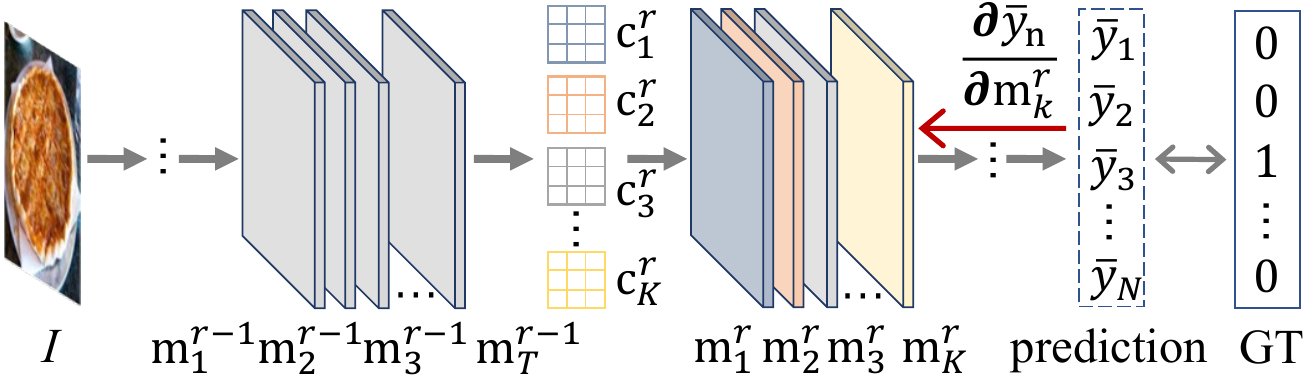}
\caption{The relationship between the predicted category and feature maps (convolutional kernels) in each layer.}
\label{fig:discovery1}
\end{figure}

\textbf{Analysis:} In this section, we first give the analysis between the convolutional kernels with the target category.
For the input image $I$, the feature maps of the $(r-1)$-th layer are denoted as $\{m_1^{r-1},m_2^{r-1},m_3^{r-1},...,m_T^{r-1}\}$, as shown in Fig. \ref{fig:discovery1}.
With the convolutional kernels $\{c^r_1,c^r_2,c^r_3,...,c^r_K\}$,
feature maps $\{m_1^r,m_2^r,m_3^r,...,m_K^r\}$ of the $r$-th layer are calculated as follows:
\begin{equation}\label{eq1}
m_k^r= f(c^r_k \circledast \{m_1^{r-1},m_2^{r-1},m_3^{r-1},...,m_T^{r-1}\} ),
\end{equation}
where $\circledast$ denotes the convolutional operation,
$f$ denotes the following operation, such as pooling operation, activation function.
With feature maps $\{m_1^r,m_2^r,m_3^r,...,m_K^r\}$ of the $r$-th layer,
the predicted category $\overline{y}_n$ is denoted as $\overline{y}_n=F(\{m_1^r,m_2^r,m_3^r,...,m_K^r\})$,
where $F$ denotes the following operations: convolution, pooling, activation function, and fully connected layer, decided by the specific network architecture.

For feature maps $\{m_1^r,m_2^r,m_3^r,...,m_K^r\}$ of the $r$-th layer,
the first-order derivative $(\overline{y}_n)'_{m_k^r}$ of the predicted category $\overline{y}_n$ w.r.t the $k$-th feature map $m_k^r$ is calculated as follows:
\begin{equation}\label{eq2}
(\overline{y}_n)'_{m_k^r} = \frac{\partial \overline{y}_n}{\partial m_k^r}.
\end{equation}
The magnitude of the derivative $(\overline{y}_n)'_{m_k^r}$ indicates which feature value in $ m_k^r$ needs to be changed the least to affect the class score $\overline{y}_n$ the most.

Then, the summed value $a^{n,r}_k$ of $k$-th first-order derivative $(\overline{y}_n)'_{m_k^r}$ is calculated as follows:
\begin{equation}\label{eq3}
a^{n,r}_k =\biguplus |\frac{\partial \overline{y}_n}{\partial m_k^r}|,
\end{equation}
where $\biguplus$ denotes summing values of the matrix $|\frac{\partial \overline{y}_n}{\partial m_k^r}|$.
The magnitude of $a^{n,r}_k$ indicates the importance of the feature map $m^r_k$ to affect the class score $\overline{y}_n$.
From Eqn. (\ref{eq1}), we can see that, with the same input feature maps $\{m_1^{r-1},m_2^{r-1},m_3^{r-1},...,m_T^{r-1}\}$ and operation $f$,
the convolutional kernel $c^r_k$ determines the different feature maps $\{m_1^r,m_2^r,m_3^r,...,m_K^r\}$.
So, the magnitude of $a^{n,r}_k$ also indicates the degree of correlation between the $k$-th convolutional kernel and the predicted category $\overline{y}_n$.

\begin{figure}[!t]
\centering
\includegraphics[scale =0.58]{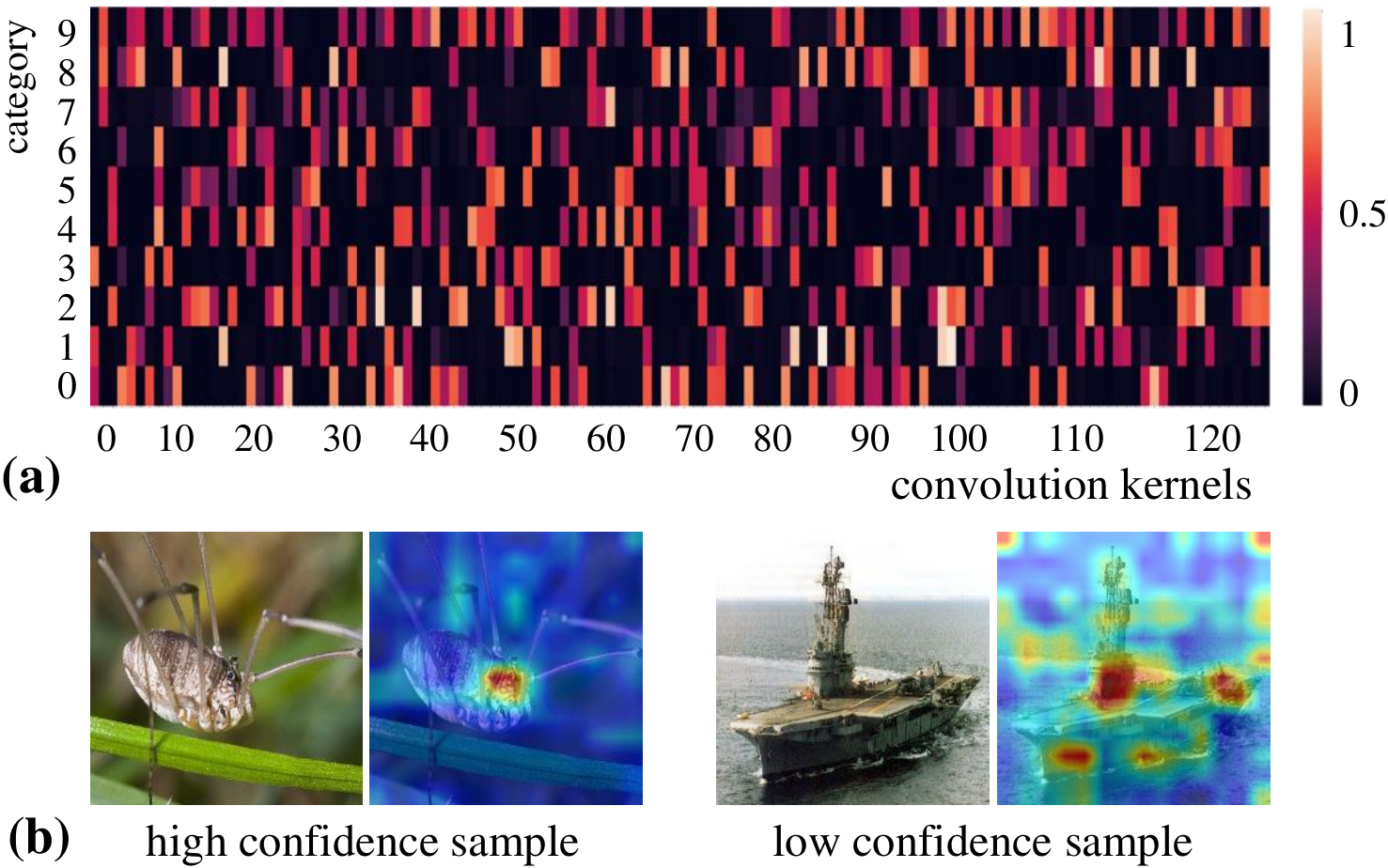}
\caption{(a) the statistical correlation diagram between all categories and the convolution kernels in last layer of GoogLeNet. (b) the high and low-confidence images covered with summed derivative maps.}
\label{fig:discovery1ab}
\end{figure}

Fig. \ref{fig:discovery1ab}(a) shows the statistical correlation degree between all categories and the convolutional kernels in the last layer of GoogLeNet \cite{szegedy2015going} on the CIFA10 dataset.
For each category, we calculate the sum value $\sum^{100}_1 a^{n,r}_k$ for each convolutional kernel $c^r_k$  of $100$ images, prediction confidence of which are larger than $0.90$.
From Fig. \ref{fig:discovery1ab}(a), we can observe that each category is only correlated with spare and specific convolutional kernels (bright color), while most convolutional kernels are not correlated (dark color).
For each image, the convolutional kernels with high correlation are almost identical with the statistical high correlation kernels in each layer of the network.
Furthermore, the number of convolutional kernels with high correlation reduces along with the network layer goes deeper.
More visual correlation diagrams about a single image and each layer are given in the \emph{supplements}.

For the image with inaccurate prediction,
the summed derivative map  $\sum^K_{k=1}|\frac{\partial \overline{y}_n}{\partial m_k^r}|$ is mapped into the original image.
Fig. \ref{fig:discovery1ab}(b) shows the high-confidence and low-confidence prediction images and the corresponding image covered with the summed derivative maps.
We can find that some background areas have a relatively high correlation. For images with accurate predictions, there are hardly any background areas with high correlations.
The above phenomenons indicate that those unrelated background features will disturb the prediction of the CNN classifier.

\begin{discovery}
Adversarial samples are isolated while normal samples are successive in the feature space.
\end{discovery}

\textbf{Analysis:} In the experiment, we find that the performance of the normal sample with disturbances is more robust than the adversarial sample.
So, we assume that adversarial samples are isolated while normal samples are successive in the feature space.
For the normal and corresponding adversarial samples obtained with fast gradient sign method \cite{goodfellow2015explaining},
 noises of different ranges are added into those samples to verify their robustness with different disturbances.
For each image, the noises are randomly added to the arbitrary layer of the network $10$ times separately.

\begin{figure}[!t]
\centering
\includegraphics[scale =0.31]{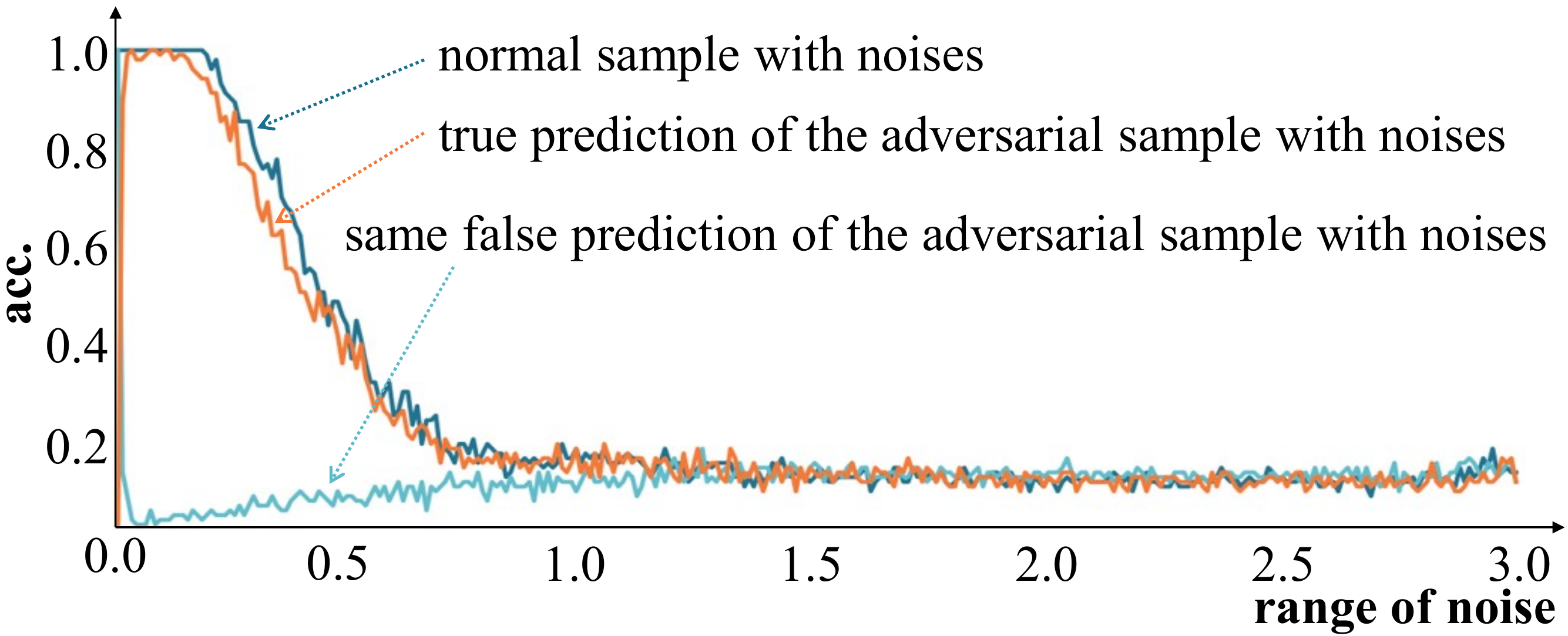}
\caption{The accuracy curves of normal and adversarial samples with different disturbances.}
\label{fig:discovery2}
\end{figure}

Fig. \ref{fig:discovery2} shows the accuracy curves of normal and adversarial samples with different disturbances.
The accuracy curves are the average results of $100$ samples.
From Fig. \ref{fig:discovery2}, we can see that a normal sample achieves accuracy close to $ 100 \%$ when the noise range in value from $0$ to $0.3$.
However, with some noises, the prediction of the adversarial sample will turn back to its' original true label and has accuracy close to $100\%$.
Only less than $10\%$ samples are still be predicted to the same false label as the adversarial sample.
So, we conclude the secondary discovery that adversarial samples are isolated while normal samples are successive in the feature space.

\subsection{Model Doctor}

Based on the above two discoveries,
we put forward a simple gradient aggregation strategy for diagnosing and treating the mainstream CNN classifiers.
In the diagnosing phase,
we accumulate the correlation between the predicted category and the convolution kernels in each layer.
In most of the case, the accurate prediction is only correlated with sparse and specific convolution kernels in the last few layers of the CNN classifier (\emph{discovery} 1).
Then, we can adopt the accumulated correlation distribution of each layer to diagnose the cause of why the sample is misclassified.
In the treating stage, two constraints are devised for treating CNN classifiers based on diagnosed results and the \emph{discovery} 2.
The channel-wise constraint is proposed to restrain the incorrect correlative convolution kernels different from the accumulated correlation distribution.
The space-wise constraint is proposed for restraining the incorrect correlation with the background features.
The details about diagnosing and treating are elaborated as follows.

\begin{figure*}[!t]
\centering
\includegraphics[scale =0.578]{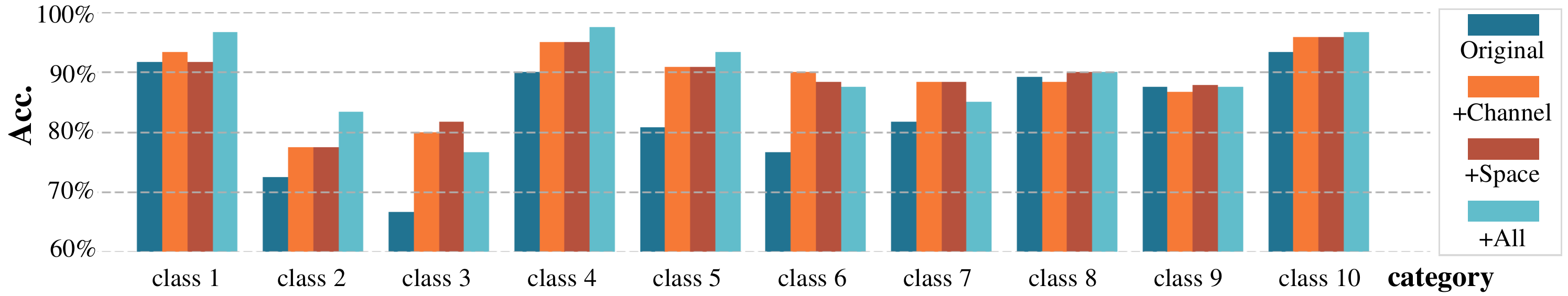}
\caption{The accuracy and increased accuracy of each category with different constraints on the mini-ImageNet dataset.}
\label{each_categoy}
\vspace{-0.5em}
\end{figure*}

\subsubsection{Diagnosing stage.}

Inspired by the \emph{discovery} 1, we accumulate the correlation between the target category and convolution kernels in each layer.
To alleviate the disturbance of misclassified features, average statistics are adopted to accumulate the correlations between the target category and the convolution kernels.
For the disturbed feature map $m_k^r+\sigma$, the noise matrix $\sigma$ sampled from the interval $[-\delta , \delta]$ will be added to the feature map $m_k^r$.
The average correlation value $\overline{a}^{n,r}_k$ between the convolution kernel  $c^r_k$ and the category $\overline{y}_n$ is calculated as follows:
\begin{equation}\label{eq4}
\overline{a}^{n,r}_k = \frac{1}{J} \sum^{J}_{j=1}  \biguplus |\frac{\partial \overline{y}_n}{\partial (m_k^r+\sigma^j)}|,
\end{equation}
where $\sigma^j$ denotes noise matrix sampled at the $j$-th time.

For convolution kernels in each layer, Eqn.(\ref{eq4}) is adopted to calculate the correlation with the target category.
For each category, $T$ samples are used for calculating the average correlation distribution,
which can illustrate the relation between the target category and the convolution kernels in each layer. $T$ is usually set to $100$ according to the experiment results.
The illustrated relationship can be used to diagnose the possible reasons for the misclassified samples.

\subsubsection{Constraint strategy.}

Before the treating stage, two constraints are devised for treating CNN classifiers based on the accumulated correlation distribution between the target category and the convolution kernels in each layer.
It's worth noting that the treating is applied to the trained classifier.
For the trained CNN classifier and training samples with high-confidence predictions, the average correlation value $\overline{a}^{n,r}_k$ between each convolution kernel $c^r_k$ and each category $\overline{y}_n$ is firstly calculated.
Then, for the training image $I$ with GT label $y_n$, the $r$-th feature maps are denoted as $\{m_1^r,m_2^r,m_3^r,...,m_K^r\}$ and the predicted label is denoted $\overline{y}_n$.
The channel-wise constraint $\mathcal{L}^r_{ch}$ on the $r$-th layer for the image $I$  is denoted as follows:
\begin{equation}\label{eq5}
\begin{aligned}
\mathcal{L}^r_{ch}= &\sum_{k=1}^{K} \mathbf{1}[\overline{a}^{n,r}_k <v]*\sum^{J}_{j=1}  \biguplus |\frac{\partial \overline{y}_n}{\partial (m_k^r+\sigma^j)}| \\
&+ \sum_{k=1}^{K} \sum^{J}_{j=1}  \biguplus |\frac{\partial \overline{y}_s}{\partial (m_k^r+\sigma^j)}|,
\end{aligned}
\end{equation}
where, the first term is used for restraining the wrong correlation between the predicted category $\overline{y}_n$ and convolution kernels of $r$-th layer that are different from average correlation,
the second term is used to restrain all the correlations between the second high-confidence prediction $\overline{y}_s$  and the convolution kernels of the $r$-th layer, $K$ denotes the number of the convolution kernel in each layer, $\sigma^j$ denotes noise matrix sampled at the $j$-th time, $\mathbf{1}[\overline{a}^{n,r}_k <v]$ equals $1$ if $\overline{a}^{n,r}_k$  is less than the threshold value $v$; otherwise equals $0$.

Furthermore, the space-wise constraint $\mathcal{L}_{sp}$ on the $r$-th layer is proposed for restraining the incorrect correlation with the background features,
which is formulated as follows:
\begin{equation}\label{eq6}
\mathcal{L}^r_{sp}=\biguplus  \sum_{k=1}^{K} \sum^{J}_{j=1}  I^r_{bg} \circledast |\frac{\partial \overline{y}_n}{\partial (m_k^r+\sigma^j)}|,
\end{equation}
where, $I^r_{bg}$ denotes the rescaled background mask that has the same size as $m_k^r$,
$\circledast$ denotes the Hadamard product. In the mask $I^r_{bg}$, the background area has value one, and the mask $I^r_{bg}$ is eroded to preserve the object boundary features.
It's worth noting that the space-wise constraint requires additional annotations, which can be rough boundaries.

\subsubsection{Treating Stage.}
In the treating stage, all constraints can be applied to any layer of the CNN classifier.
Based on the fact that deep layers of the CNN classifier usually contain high-level semantic features,
the space-wise and channel-wise constraints are adopted to constrain the deep layers. The following ablation study on layer depth also verifies the practicability and effectiveness of the above constraining way.
Furthermore, the above two constraints are usually appended to the original training loss $\mathcal{L}_{orig}$ as follows:
\begin{equation}\label{eq7}
\mathcal{L}_{all}=\mathcal{L}_{orig}+\sum_{r \in \mathbb{S}_{sp}} \mathcal{L}^r_{sp}+\sum_{r \in \mathbb{S}_{ch}} \mathcal{L}^r_{ch},
\end{equation}
where, $\mathbb{S}_{sp}$ and $\mathbb{S}_{ch}$ denote the constrained layer sets of space-wise constraint and channel-wise constraint, respectively.
 The two constraints also could be appended to the original training loss separately.
 In the treating stage, the parameter setting and the optimizer are the same as the setting of the original CNN classifier.

\begin{table*}[!t]
\footnotesize
\centering
\resizebox{\textwidth}{!}{
\begin{tabular}{ccccccccc}
\toprule
\textbf{Dataset} &\textbf{MNIST} & \textbf{Fashion-MNIST} &\textbf{CIFAR-10} &\textbf{CIFAR-100} & \textbf{SVHN} & \textbf{STL-10} &\textbf{mini-ImageNet}  \\
\midrule
\textbf{AlexNet / +All} &99.60 /\ $\quad$--$\quad$  &93.32 /\ $\quad$--$\quad$           &86.32 /\ $\quad$--$\quad$     &55.04 /\ $\quad$--$\quad$       &93.44 /\ $\quad$--$\quad$   &67.59 /\ +0.66     &76.92 /\ +3.84    \\
\textbf{+Space / +Channel}    &-1.63 /\ +0.01&-1.44 /\ +0.09    &-1.91 /\ +0.31&-6.81 /\ +1.87    &-1.65 /\ +0.08&+0.59 /\ +0.62 &+3.74 /\ +3.21   \\
\midrule
\textbf{VGG-16 / +All}  &99.71 /\ $\quad$--$\quad$  &95.21 /\ $\quad$--$\quad$           &93.46 /\ $\quad$--$\quad$     &70.39 /\ $\quad$--$\quad$       &94.72 /\ $\quad$--$\quad$   &77.65 /\ +0.79     &83.00 /\ +5.92     \\
\textbf{+Space / +Channel}    &-0.98 /\ +0.03&-1.91 /\ +0.10    &-2.52 /\ +0.74&-7.92 /\ +1.08    &-1.67 /\ +0.07&+0.69 /\ +0.73 &+5.58 /\ +5.50  \\
\midrule
\textbf{ResNet-50 / +All}  &99.73 /\ $\quad$--$\quad$  &95.33 /\ $\quad$--$\quad$           &94.85 /\ $\quad$--$\quad$     &77.08 /\ $\quad$--$\quad$       &94.81 /\ $\quad$--$\quad$   &82.14 /\ +0.63     &90.25 /\ +3.71     \\
\textbf{+Space / +Channel}    &-1.20 /\ +0.00&-0.61 /\ +0.15    &-3.44 /\ +0.54&-4.70 /\ +0.95    &-0.92 /\ +0.04&+0.61 /\ +0.58 &+3.61 /\ +3.59   \\
\midrule
\textbf{SENet-34 / +All}  &99.75 /\ $\quad$--$\quad$  &95.35 /\ $\quad$--$\quad$           &94.76 /\ $\quad$--$\quad$     &74.95 /\ $\quad$--$\quad$       &94.67 /\ $\quad$--$\quad$   &81.67 /\ +0.92     &89.23 /\ +2.73     \\
\textbf{+Space / +Channel}    &-0.88 /\ +0.01&-3.63 /\ +0.15    &-1.89 /\ +0.41&-6.21 /\ +1.22    &-1.30 /\ +0.08&+0.87 /\ +0.73 &+2.12 /\ +2.63   \\
\midrule
\textbf{WideResNet-28 / +All}  &99.47 /\ $\quad$--$\quad$  &93.81 /\ $\quad$--$\quad$           &94.26 /\ $\quad$--$\quad$     &77.48 /\ $\quad$--$\quad$       &94.11 /\ $\quad$--$\quad$   &79.34 /\ +0.85     &88.47 /\ +3.26      \\
\textbf{+Space / +Channel}    &-1.72 /\ +0.19&-3.70 /\ +1.45    &-2.94 /\ +0.35&-5.01 /\ +0.39    &-2.30 /\ +0.02&+0.82 /\ +0.81 &+3.24 /\ +3.15   \\
\midrule
\textbf{ResNeXt-50 / +All}  &99.69 /\ $\quad$--$\quad$  &95.37 /\ $\quad$--$\quad$           &94.34 /\ $\quad$--$\quad$     &74.76 /\ $\quad$--$\quad$       &94.25 /\ $\quad$--$\quad$   &83.21 /\ +0.49     &89.72 /\ +3.42   \\
\textbf{+Space / +Channel}    &-1.34 /\ +0.01&-2.01 /\ +0.19    &-1.77 /\ +1.21&-6.05 /\ +2.26    &-1.81 /\ +0.03&+0.34 /\ +0.47 &+3.19 /\ +3.21  \\
\midrule
\textbf{DenseNet-121 / +All}  &99.72 /\ $\quad$--$\quad$  &95.43 /\ $\quad$--$\quad$           &95.22 /\ $\quad$--$\quad$     &76.92 /\ $\quad$--$\quad$       &95.18 /\ $\quad$--$\quad$   &84.03 /\ +0.91     &89.83 /\ +3.29     \\
\textbf{+Space / +Channel}    &-0.70 /\ +0.01&-3.38 /\ +0.00    &-3.81 /\ +0.52&-5.79 /\ +0.75    &-0.94 /\ +0.01&+0.84 /\ +0.69 &+2.98 /\ +2.97   \\
\midrule
\textbf{SimpleNet-v1 / +All}  &99.72 /\ $\quad$--$\quad$  &95.39 /\ $\quad$--$\quad$           &94.61 /\ $\quad$--$\quad$     &75.29 /\ $\quad$--$\quad$       &94.51 /\ $\quad$--$\quad$   &81.92 /\ +0.67     &87.92 /\ +2.19      \\
\textbf{+Space / +Channel}    &-1.28 /\ +0.01&-2.75 /\ +0.11    &-3.67 /\ +0.80&-4.93 /\ +1.51    &-1.36 /\ +0.07&+0.66 /\ +0.53 &+1.97 /\ +1.82   \\
\midrule
\textbf{EfficientNetV2-S / +All}  &99.66 /\ $\quad$--$\quad$  &93.82 /\ $\quad$--$\quad$           &91.07 /\ $\quad$--$\quad$     &61.01 /\ $\quad$--$\quad$       &92.14 /\ $\quad$--$\quad$   &79.32 /\ +0.78     &86.33 /\ +4.84     \\
\textbf{+Space / +Channel}    &-1.11 /\ +0.05&-1.21 /\ +0.08    &-1.09 /\ +0.33&-3.91 /\ +2.65    &-1.55 /\ +0.05&+0.55 /\ +0.71 &+4.73 /\ +4.67   \\
\midrule
\textbf{GoogLeNet / +All}  &99.75 /\ $\quad$--$\quad$  &95.20 /\ $\quad$--$\quad$           &94.36 /\ $\quad$--$\quad$     &75.28 /\ $\quad$--$\quad$       &94.20 /\ $\quad$--$\quad$   &83.33 /\ +0.93     &91.33 /\ +2.32      \\
\textbf{+Space / +Channel}    &-0.90 /\ +0.01&-2.98 /\ +0.17    &-4.67 /\ +0.58&-6.34 /\ +0.71    &-0.82 /\ +0.13&+0.63 /\ +0.92 &+1.95 /\ +2.21   \\
\midrule
\textbf{Xception / +All}  &99.75 /\ $\quad$--$\quad$  &95.23 /\ $\quad$--$\quad$           &93.35 /\ $\quad$--$\quad$     &74.15 /\ $\quad$--$\quad$       &92.38 /\ $\quad$--$\quad$   &79.62 /\ +0.96     &84.25 /\ +3.81     \\
\textbf{+Space / +Channel}    &-1.04 /\ +0.02&-1.40 /\ +0.05    &-3.81 /\ +0.57&-5.45 /\ +0.42    &-1.20 /\ +0.04&+0.84 /\ +0.93 &+4.52 /\ +4.72   \\
\midrule
\textbf{MobileNetV2 / +All}  &99.68 /\ $\quad$--$\quad$  &94.58 /\ $\quad$--$\quad$           &91.95 /\ $\quad$--$\quad$     &68.26 /\ $\quad$--$\quad$       &90.43 /\ $\quad$--$\quad$   &78.92 /\ +0.67     &86.21 /\ +4.79      \\
\textbf{+Space / +Channel}    &-2.83 /\ +0.00&-2.91 /\ +0.39    &-2.08 /\ +1.07&-7.91 /\ +2.75    &-0.73 /\ +0.08&+0.65 /\ +0.52 &+4.61 /\ +4.51  \\
\midrule
\textbf{Inception-v3 / +All}  &99.70 /\ $\quad$--$\quad$  &95.25 /\ $\quad$--$\quad$           &94.77 /\ $\quad$--$\quad$     &76.93 /\ $\quad$--$\quad$       &93.54 /\ $\quad$--$\quad$   &89.46 /\ +0.78     &89.41 /\ +4.10     \\
\textbf{+Space / +Channel}    &-1.20 /\ +0.06&+1.29 /\ +0.08    &-2.56 /\ +0.52&-7.24 /\ +1.20    &-1.72 /\ +0.01&+0.72 /\ +0.69 &+3.83 /\ +3.70   \\
\midrule
\textbf{ShuffleNetV2 / +All}  &99.69 /\ $\quad$--$\quad$  &94.63 /\ $\quad$--$\quad$           &92.28 /\ $\quad$--$\quad$     &67.39 /\ $\quad$--$\quad$       &92.47 /\ $\quad$--$\quad$   &75.21 /\ +0.64     &87.08 /\ +2.69      \\
\textbf{+Space / +Channel}    &-1.01 /\ +0.00&-3.62 /\ +0.19    &-2.81 /\ +0.64&-3.23 /\ +0.43    &-1.26 /\ +0.06&+0.56 /\ +0.43 &+2.60 /\ +2.57   \\
\midrule
\textbf{SqueezeNet / +All}  &99.71 /\ $\quad$--$\quad$  &94.74 /\ $\quad$--$\quad$           &91.93 /\ $\quad$--$\quad$     &67.87 /\ $\quad$--$\quad$       &92.65 /\ $\quad$--$\quad$   &78.75 /\ +0.59     &88.58 /\ +3.52      \\
\textbf{+Space / +Channel}    &-0.92 /\ +0.01&-1.45 /\ +0.18    &-2.76 /\ +0.92&-6.81 /\ +0.60    &-0.98 /\ +0.06&+0.57 /\ +0.59 &+3.19 /\ +3.01  \\
\midrule
\textbf{MnasNet / +All}  &99.66 /\ $\quad$--$\quad$  &93.53 /\ $\quad$--$\quad$           &85.55 /\ $\quad$--$\quad$     &53.60 /\ $\quad$--$\quad$       &91.08 /\ $\quad$--$\quad$   &74.32 /\ +0.31     &87.33 /\ +3.49      \\
\textbf{+Space / +Channel}    &-1.14 /\ +0.01&-2.53 /\ +0.03    &-2.58 /\ +1.37&-3.90 /\ +0.50    &-1.04 /\ +0.03&+0.29 /\ +0.19 &+2.08 /\ +3.27   \\
\bottomrule
\end{tabular}}
\vspace{-1em}
\caption{The base and improved accuracy of $16$ classifiers on $7$ datasets.
For the small-size image, due to the expanded receptive field of convolution operation, the feature map of the last few layers will contain mixture features of object and backgrounds, which leads to negative effective of  the space-wise constraint. So, ``+All'' are not applied to the small-size image datasets.}
\label{comparing_SOTA}
\vspace{-0.5em}
\end{table*}

\section{Experiments}

In the experiment, the adopted classifiers, datasets, and experiment settings are listed as follows.

\textbf{Classifier.}
The selected $16$ classifiers cover mainstream classification network architectures, which are listed as follows:
AlexNet~\cite{2012ImageNet},
VGG-16~\cite{2014Very},
ResNet-34~\cite{2016Deep},
ResNet-50~\cite{2016Deep},
WideResNet-28~\cite{zagoruyko2016wide},
ResNeXt-50~\cite{xie2017aggregated},
DenseNet-121~\cite{2017Densely},
SimpleNet-v1~\cite{hasanpour2016lets},
EfficientNetV2-S~\cite{tan2021efficientnetv2},
GoogLeNet~\cite{2014Going},
Xception~\cite{chollet2017xception},
MobileNetV2~\cite{2017MobileNets},
Inception-v3~\cite{szegedy2016rethinking},
ShuffleNetV2~\cite{2017ShuffleNet},
SqueezeNet~\cite{2016SqueezeNet:},
and MnasNet~\cite{tan2019mnasnet}.

\textbf{Dataset.}
The datasets we adopted contain
MNIST~\cite{lecun2001gradient},
Fashion-MNIST~\cite{xiao2017fashion},
CIFAR-10,
CIFAR-100~\cite{krizhevsky2009learning},
SVHN~\cite{netzer2011reading},
STL-10~\cite{coates2011an} and
mini-ImageNet~\cite{vinyals2016matching}, which are commonly used datasets for the classification task.

\textbf{Experiment setting.} Unless stated otherwise, the default experiment settings are given as follows: $J=10$, $T=100$,
the expanded pixel number is $30$.
More experiment details are given in the \emph{supplements}.

\subsection{The Effect of Model Doctor for SOTA Classifiers}
In this section, we conduct massive experiments of $16$ mainstream CNN classifiers on $7$ datasets. All results are averages of three runs.
For the space-wise constraint, we only annotated $20$ low-confidence images for each category.
Of course, the classifier can achieve higher accuracy with more annotations.
Before diagnose and treating, all classifiers have achieved the optimal classification performance in the setting of original training.
From Table \ref{comparing_SOTA}, we can see that Model Doctor helps all CNN classifiers improve accuracy by $1\%\sim 5\%$, which verifies the versatility and effectiveness of the proposed gradient aggregation strategy.
For datasets with small-size images, only the channel-wise constraint can improve the accuracy.
On the contrary, the space-wise constraint has negative effects on those small-size image datasets.
The root reason is that the expanded receptive field of convolution operation brings the mixture features of objects and backgrounds in the feature map of the last few layers, which leads to the wrong scope constraint of the small-size image.
What's more, Model Doctor has little effect on high-performance networks such as, $99.7\%$ accuracy on MNIST dataset.
For the regular size image datasets (STL-10 and mini-ImageNet), the space-wise constraint achieves a significant accuracy increase,
which indicates that background area features indeed have some distractions for classifiers.
Model Doctor only achieves accuracy close to $1\%$ on STL-10. The possible reason is that the object almost takes up the whole image.

To verify the effectiveness of Model Doctor on various categories, we conduct experiments on randomly selected $10$ categories  $10$ times for the mini-ImageNet dataset, which contains $100$ categories and $480$ samples for each category. Table \ref{comparing_SOTA} shows the average results of $10$ times.
More experiments on mini-ImageNet are given in the \emph{supplements}.
Fig. \ref{each_categoy} shows the accuracy and increased accuracy of VGG16 on a $10$ category classification task.
We can see that Model Doctor can improve higher accuracy increase on low accuracy categories, which indicates that Model Doctor can accurately diagnose the classifier failures and effectively treat them.

\subsection{Visual Results}

\begin{figure}[!t]
\centering
\includegraphics[scale =0.53]{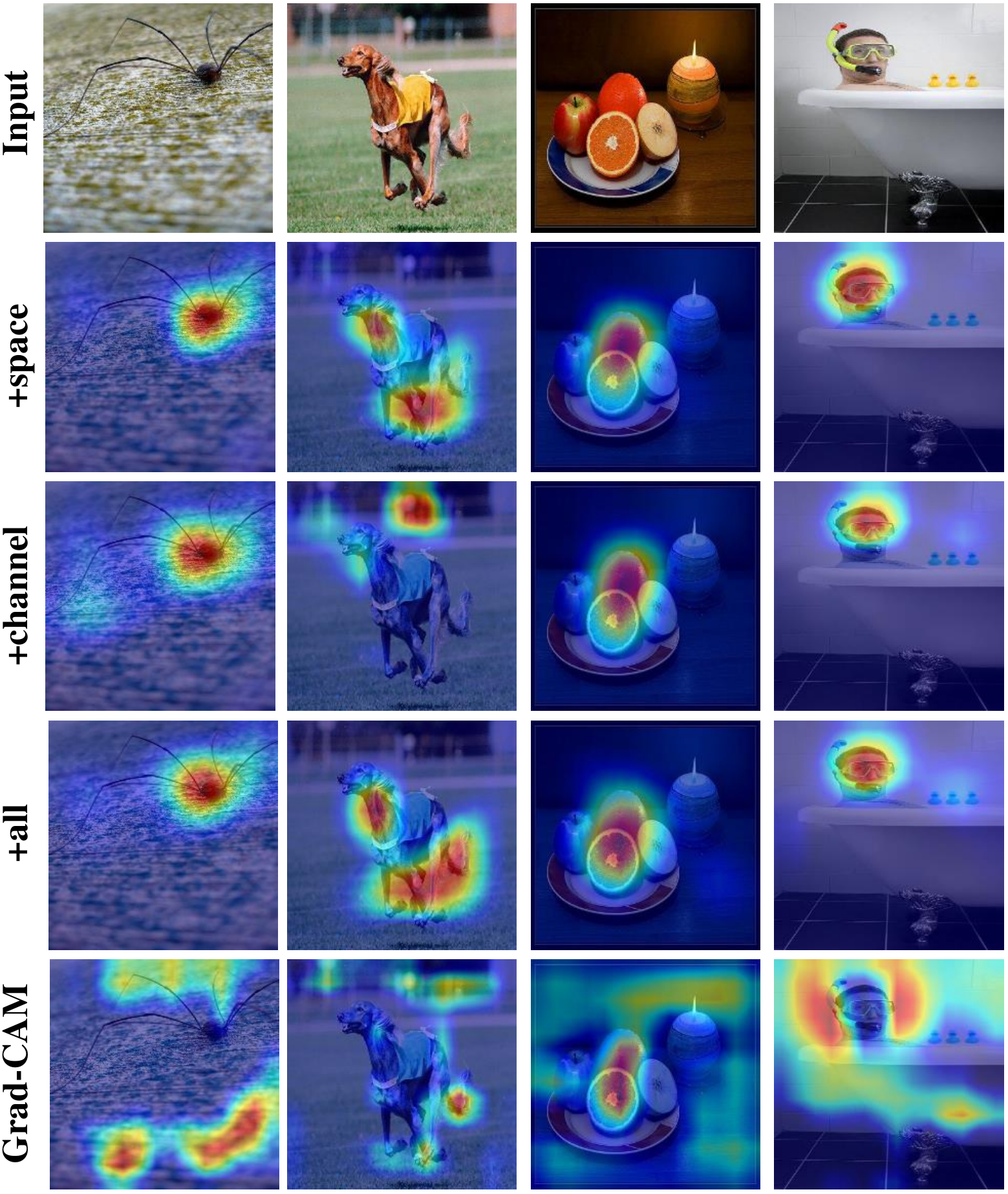}
\caption{Visual results of original images with different constraints and Grad-CAM on mini-ImageNet datasets.}
\label{visualresults}
\vspace{-0.5em}
\end{figure}

Furthermore, Fig. \ref{visualresults} shows the visual results of
original images with different constraints and Grad-CAM for visualizing the differences intuitively.
We can see that the space-wise constraint successfully restrains the irrelevant background areas.
For the visual results of channel-wise constraint, most irrelevant areas are also restrained.
The association maps calculated with Grad-CAM contain some background areas. The proposed Model Doctor not only can improve the classification accuracy but also regularize the association between the category and the input features.

\subsection{The Ablation Study}

\begin{figure}[!t]
\centering
\includegraphics[scale =0.30]{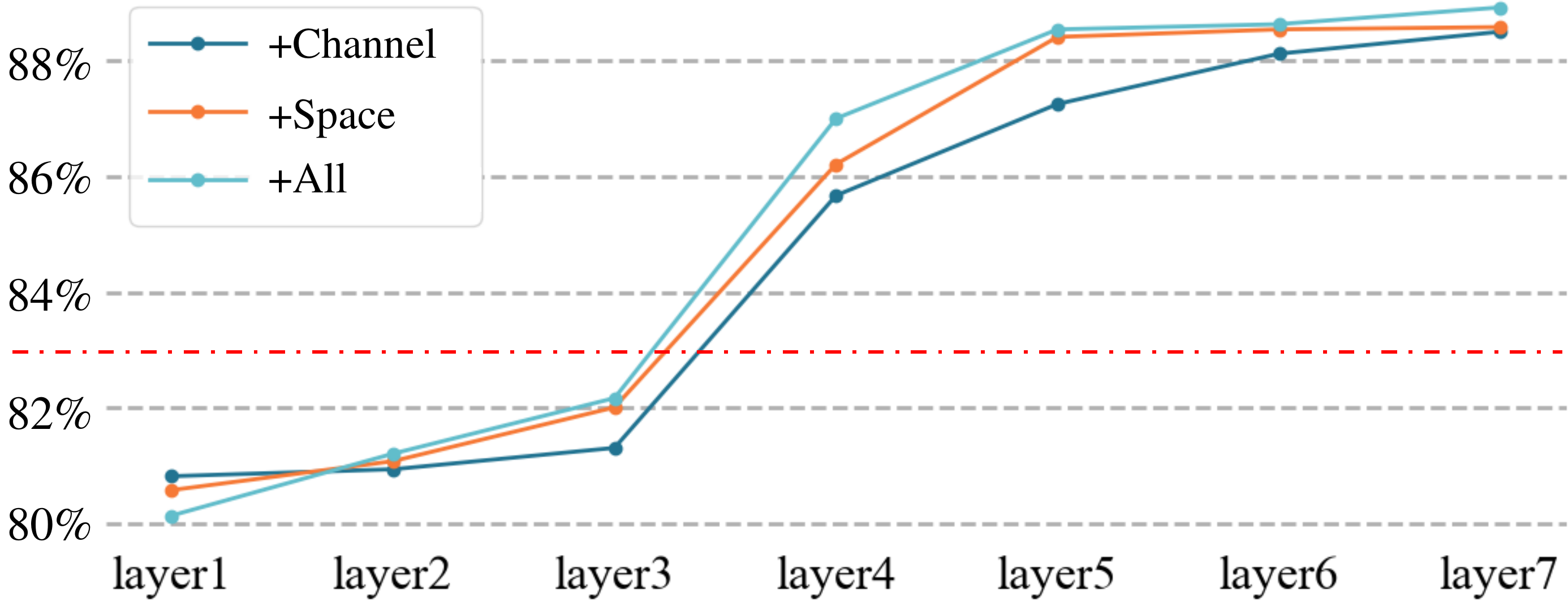}
\caption{The ablation study of constraints on layer depth. Red dash line denotes the original accuracy ($83\%$).}
\label{fig:ablation_study_channel}
\end{figure}

This section conducts the ablation study of channel-wise constraint and space-wise constraint on mini-ImageNet with VGG16.
The \emph{layer depth impact} for different constraints is shown in Fig. \ref{fig:ablation_study_channel}, where
we can see that the increased accuracy of all constraints will increase when the layer goes deep.
It reveals that the proposed Model Doctor is more suitable for deep layers of classifiers.
The possible reason is that the deep and shallow layers contain more high-level semantic features and low-level features, respectively.
Constraining shallow layer will disturb feature extraction ability of classifier on essential features.

Furthermore, we conduct the ablation study on the \emph{annotated sample number} and the \emph{expanded pixel number} of the background mask. From Table \ref{annotation}, we can see that $12\%$ annotated samples for space-wise constraint have achieved near-optimal accuracy increase,
which indicates that only small part annotations are enough for the space-wise constraint.
Table \ref{pixel} shows that $30$-pixel expansion achieves the best performance and the expanded mask with less than $5$ pixels achieves a negative effect, the reason of which is that boundary features are crucial for the classification task.

\subsection{Discussion and Future Work}

From Table \ref{comparing_SOTA} , we can see that the treated classifier still can't achieve $100\%$ accuracy.
The possible reasons contain multiple aspects:
1) hard samples have different correlations with convolution kernels from the correlations of normal samples;
2) the training samples are not sufficient for the classifier;
3) the classifier may be over-/under-fitting. We will explore actual causes in future work.

In this paper, we verify Model Doctor on mainstream CNN classifiers. Actually, Model Doctor can be extended to a unified framework for diagnosing and treating deep models for different tasks.
We will focus on exploring the extension of Model Doctor on segmentation, detection, and key points identification tasks in the future.

\begin{table}[!t]
\footnotesize
\centering
\resizebox{0.48\textwidth}{!}{
\begin{tabular}{ccccccccc}
\toprule
& \textbf{Number} & 100 & 200 & 400 & 600 & 800  & 1000   \\
\emph{VGG16} & \textbf{Ratio}  & 2$\%$ &4$\%$ & 8$\%$ & 12$\%$  &16$\%$  &20$\%$   \\
 \cmidrule(r){2-8}
83.00\% & \textbf{+Space} &+3.22\%  &+5.58\%   &+6.28\%    &+7.53\% &+7.96\%   &+8.03\%     \\

\bottomrule
\end{tabular}}
\caption{The ablation study of space-wise constraint on annotation samples.}
\label{annotation}
\end{table}

\begin{table}[!t]
\footnotesize
\centering
\resizebox{0.48\textwidth}{!}{
\begin{tabular}{cccccccc}
\toprule
\emph{VGG16} & \textbf{Pixel}  & 0  & 5  & 10  & 20  & 30   & 40     \\
 \cmidrule(r){2-8}
83.00\% & \textbf{+Space} &-7.48\% &+0.43\%  &+3.45\%   &+4.23\%   &+5.58\% &+4.07\%     \\
\bottomrule
\end{tabular}}
\caption{The ablation study of space-wise constraint on mask expansion.}
\label{pixel}
\vspace{-0.5em}
\end{table}

\section{Conclusion}
In this paper, we put forward a universal Model Doctor for diagnosing and treating CNN classifiers in a completely automatic manner.
Firstly, we explore and validate two discoveries that 1) each category is only correlated with sparse and specific convolution kernels, and 2) adversarial samples are isolated while normal samples are successive in the feature space,
which can serve as grounds for future researches.
Based on the above two discoveries, a simple gradient aggregation strategy is devised for effectively diagnosing and optimizing CNN classifiers. Massive experiments reveal that the devised two constraints are suitable for deep layers of the classifier.
For trained classifiers, the proposed Model Doctor can effectively help them improve accuracy by $1\%\sim5\%$.
Model Doctor can be used as a convenient tool for researchers to optimize their CNN classifiers.

\section{ Acknowledgments}
This work is funded by the National Key R\&D Program of China (Grant No: 2018AAA0101503) and the Science and technology project of SGCC (State Grid Corporation of China): fundamental theory of human-in-the-loop hybrid-augmented intelligence for power grid dispatch and control.

\bibliography{AAAI}

\newpage

\textbf{----------------- Supplementary Materials -----------------}

In the supplementary materials, we provide experiment details, more visual results, and experiment analysis.
\subsection*{A. Experiment Setting}

The detail and expanded pixels for each dataset are given in Table \ref{dataset}.
The depths (convolution layers) for $16$ networks are listed as follows:
AlexNet: 5,
VGG-16: 13,
ResNet-50: 53,
SENet-34: 36,
WideResNet-28: 28,
ResNeXt-50: 53,
DenseNet-121: 120,
SimpleNet-v1: 13,
EfficientNetV2-S: 112,
GoogLeNet: 66,
Xception: 74,
MobileNetV2: 54,
Inception-v3: 94,
ShuffleNetV2: 56,
SqueezeNet: 26,
and MnasNet: 52.
For a fair comparison of different networks, the $\mathbb{S}_{sp}$ (constrained layer sets of space-wise constraint) and $\mathbb{S}_{ch}$ (constrained layer sets of channel-wise constraint) are set as the last two layers and the last one layer, respectively.
For different networks, it will achieve more accuracy increases selecting appropriate constraining layers.
Empirically, constraining the deeper layers can achieve better performance, as shown in the ablation study (Fig. 6 of the submmitted paper.)

\subsection*{B. The Correlation Between Categories and Convolution Kernels}
For demonstrating the universality of \emph{Discovery 1 that the target category is only correlated with sparse and specific convolutional kernels in the last few layers of the CNN classifier},
we provide more statistical correlation diagrams of different networks on CIFAR-10 dataset.
From Fig. \ref{digram16}, we can see that correlation between the target category and convolution kernels are sparse for all networks,
which verify the generality and universality of \emph{Discovery 1}.

\begin{figure}[!t]
\centering
\includegraphics[scale =0.58]{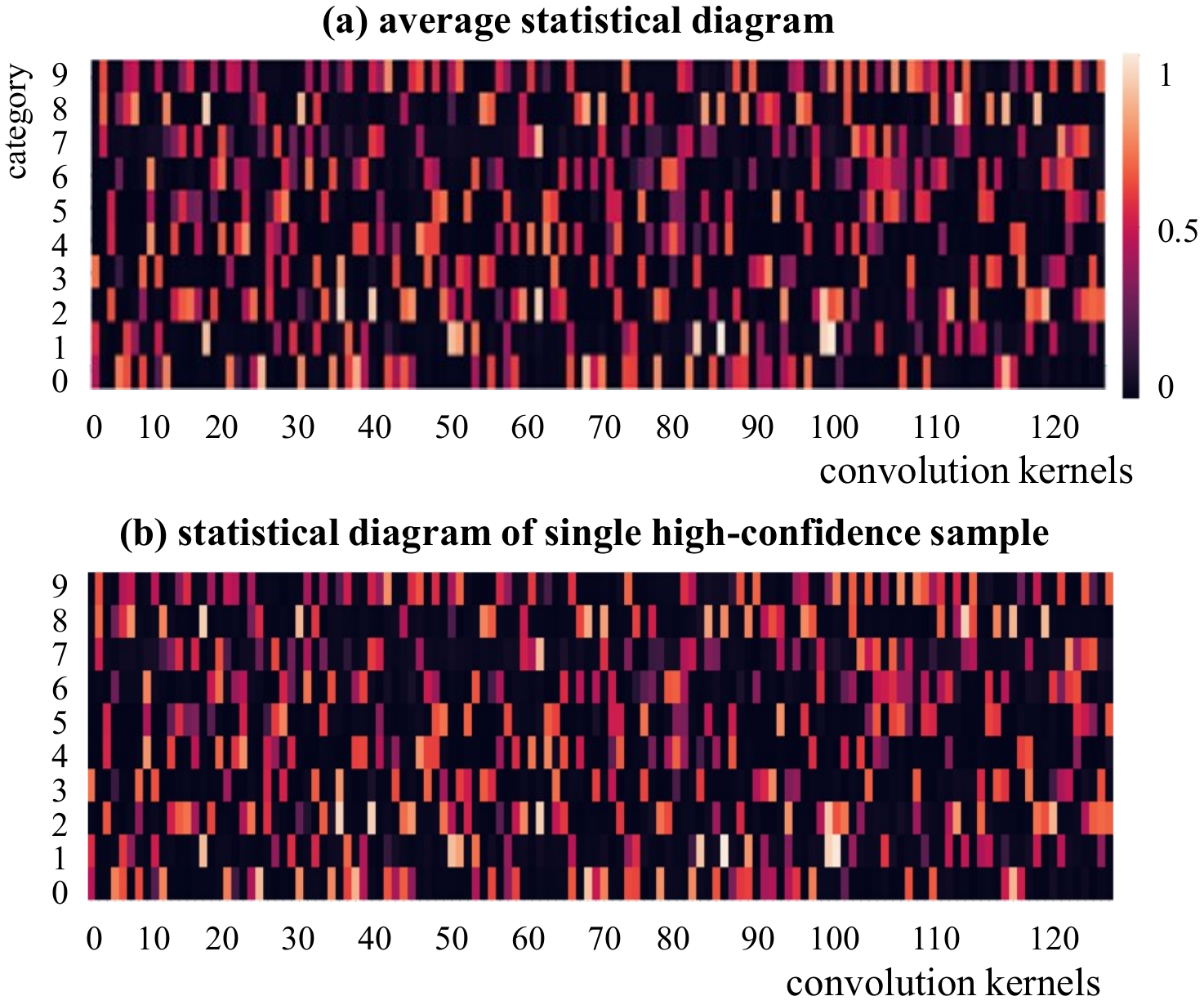}
\caption{The comparison between the (a) average statistical diagram and (b) statistical diagram of a single high-confidence sample.
The similar statistical diagrams demonstrate that the correlation between category and convolution kernels is almost same for high-confidence samples of the same category.  }
\label{averageVSsingle}
\end{figure}

Fig. \ref{averageVSsingle} shows the comparison between the average statistical diagram and statistical diagram of a single high-confidence sample.
From Fig. \ref{averageVSsingle}, we can observe that two statistical diagrams are almost the same,
which indicates that the correlation between category and convolution kernels is almost the same for high-confidence samples of the same category.
It also verifies that the target category is only correlated with specific convolutional kernels in each layer of the classifier.

\begin{figure}[!t]
\centering
\includegraphics[scale =0.95]{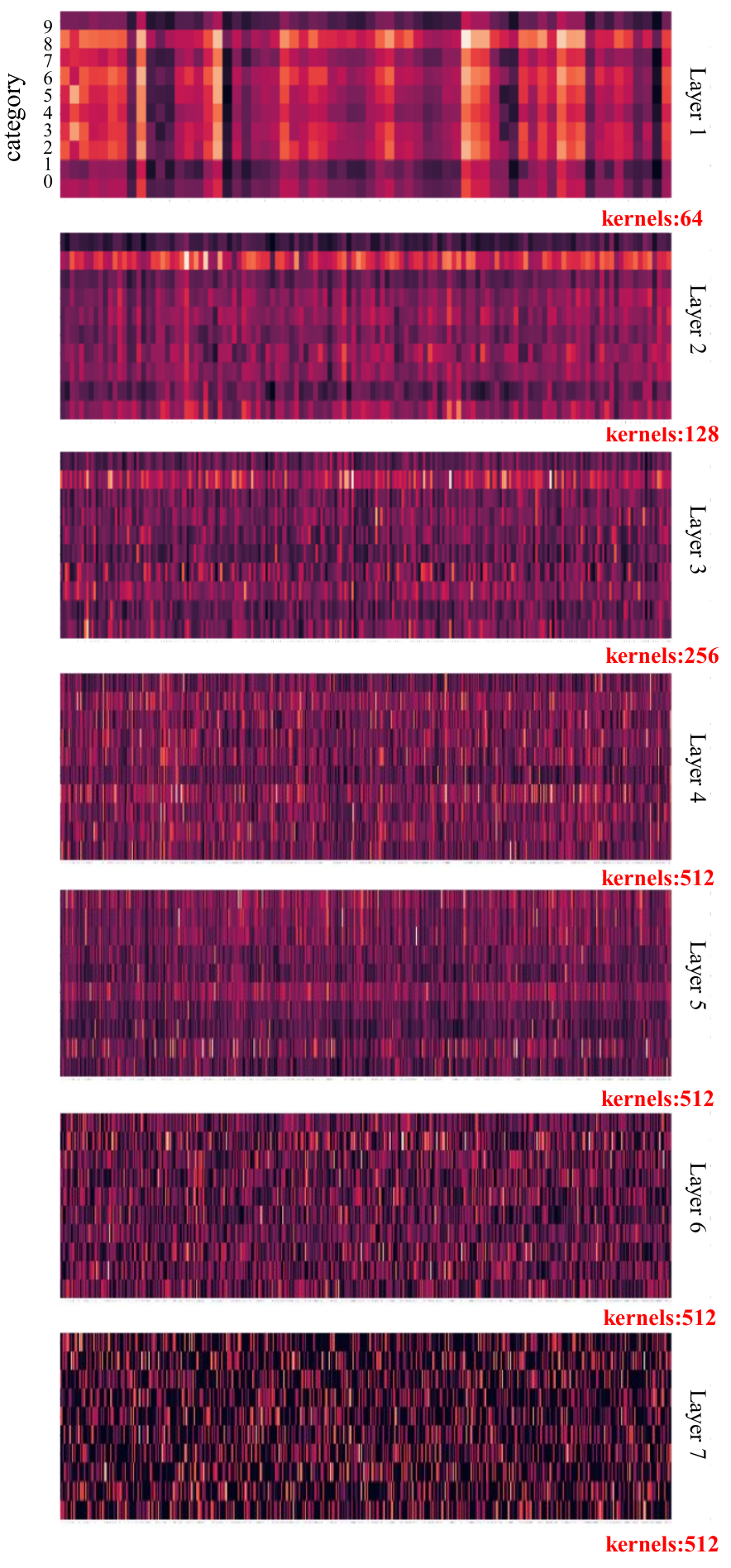}
\caption{The correlation diagrams  of different layers in GoogLeNet.  }
\label{DifferentLayer}
\end{figure}

\begin{figure*}[!t]
\centering
\includegraphics[scale =0.98]{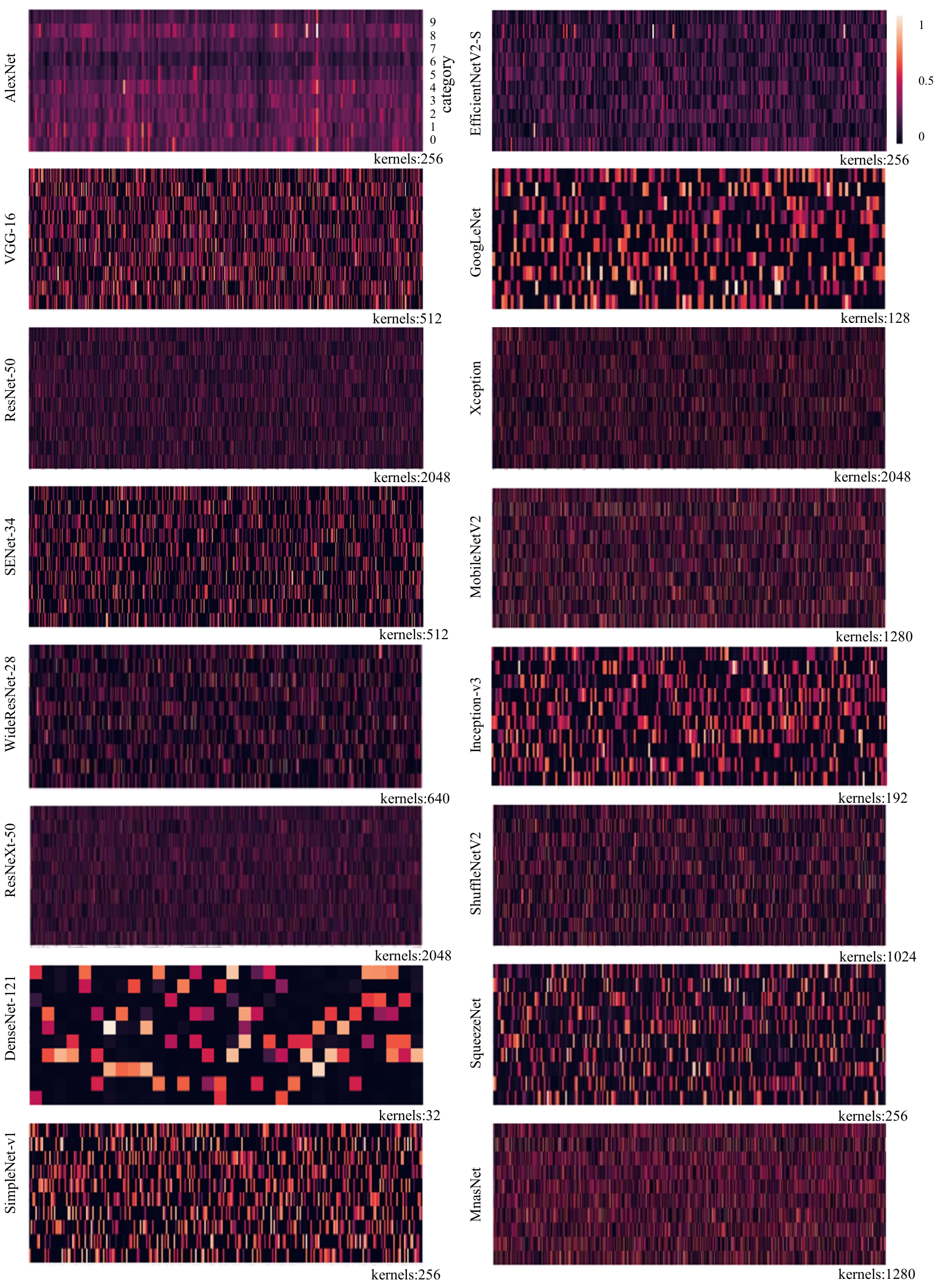}
\caption{The average statistical diagram of different networks on CIFAR-10 dataset.  }
\label{digram16}
\end{figure*}

What's more, we provide the correlation diagram of different layers in GoogLeNet, which is shown in Fig. \ref{DifferentLayer}.
As the layer goes deeper, the correlation becomes more sparse, which is consistent with common sense that the shallow and deep layers of the network contain low-level basic features and high-level semantic features, respectively.

\begin{table}[!t]
\footnotesize
\centering
\resizebox{0.46\textwidth}{!}{
\begin{tabular}{cccc}
\toprule

\textbf{Dataset}       & \textbf{Image size} & \textbf{Category}  & \textbf{Expansion} \\
\midrule
\textbf{MNIST}         & 28$\times$28        &  10                              &  2\\
\textbf{Fashion-MNIST} & 28$\times$28        &  10                              &  2\\
\textbf{CIFAR-10}      & 32$\times$32        &  10                                &  3\\
\textbf{CIFAR-100}     & 32$\times$32        &  100                             &  3\\
\textbf{SVHN}          & 32$\times$32        &  10                               &  3\\
\textbf{STL-10}        & 96$\times$96        &  10                           &  10\\
\textbf{mini-ImageNet} & 224$\times$224      &  100                            &  30\\
\midrule
\bottomrule
\end{tabular}}
\caption{Dataset details and settings. $\mathbb{S}_{sp}$ and $\mathbb{S}_{ch}$ denote the constrained layer sets of space-wise constraint and channel-wise constraint, respectively. `Expansion' denotes the expanded pixel number of the background mask.}
\label{dataset}
\end{table}

\subsection*{C. Experiment Results and Analysis }

In this section, we give the experiments on mini-ImageNet with 100 categroies (Table \ref{miniImageNet100}), and more analysis about the base and improved accuracy of 16 classifiers on 7 datasets (Table 1 of the submitted paper). Model Doctor achieves significant classification performance improvement on min-ImageNet.
For the small-size image datasets (MNIST, Fashion-MNIST, CIFAR-10, CIFAR-100, and SVHN),
Model Doctor achieves $0.01\% \sim 2.65\%$ increase.
The fundamental reasons contain two parts:
1) the image is too small, which leads to a few background feature disturbances.
2) the simple image components bring high accuracy score close to $100\%$, which leaves less space for improvement.
For the STL-10, Model Doctor only helps classifiers improve about $0.2\% \sim 1\%$ accuracy,
which is caused by the fact that the target object occupied most image areas. There are fewer disturbing background features.
For the classification task on CIFAR-100, Model Doctor can achieve $0.4\% \sim 2.7\%$ improvements.
In summary, Model Doctor is suitable for challenging classification task where the sample contains complex background and has standard image size.

\begin{table}[!t]
\footnotesize
\centering
\resizebox{0.46\textwidth}{!}{
\begin{tabular}{ccccc}
\toprule
 &\multicolumn{3}{c}{\textbf{mini-ImageNet (100 categories)} }\\
 \cmidrule(r){2-5}
\textbf{Network}& Base & +Space & +Channel & +All\\
\midrule
\textbf{AlexNet} &60.95 &+4.83 &+3.51     &+4.89 \\
\midrule
\textbf{VGG-16} &77.44 &+5.27 &+4.75     &+5.42 \\
\midrule
\textbf{ResNet-50} &81.35 &+4.29 &+3.88     &+4.30 \\
\midrule
\textbf{SENet-34} &80.27 &+4.62 &+4.03     &+4.74 \\
\midrule
\textbf{WideResNet-28} &80.31 &+4.60 &+4.17     &+4.62 \\
\midrule
\textbf{ResNeXt-50} &81.11 &+5.12 &+4.89     &+5.23 \\
\midrule
\textbf{DenseNet-121} &82.32 &+3.19 &+3.28     &+3.51 \\
\midrule
\textbf{SimpleNet-v1} &79.18 &+3.22 &+3.19     &+3.24 \\
\midrule
\textbf{EfficientNetV2-S} &83.30 &+4.98 &+4.52     &+5.12 \\
\midrule
\textbf{GoogLeNet} &82.83 &+4.44 &+4.36     &+4.50 \\
\midrule
\textbf{Xception} &83.23 &+2.19 &+3.84     &+3.99 \\
\midrule
\textbf{MobileNetV2} &83.18 &+3.67 &+3.28     &+3.71 \\
\midrule
\textbf{Inception-v3} &82.58 &+4.01 &+3.91     &+4.16 \\
\midrule
\textbf{ShuffleNetV2} &81.98 &+3.82 &+3.87     &+3.89 \\
\midrule
\textbf{SqueezeNet} &74.48 &+4.10 &+2.60     &+4.11 \\
\midrule
\textbf{MnasNet} &74.90 &+3.89 &+3.01     &+4.05 \\
\bottomrule
\end{tabular}}
\caption{The base and improved accuracy of $16$ classifiers on mini-ImageNet with 100 categroies.}
\label{miniImageNet100}
\end{table}

\subsection*{D. Visual Result}
\begin{figure*}[!t]
\centering
\includegraphics[scale =0.95]{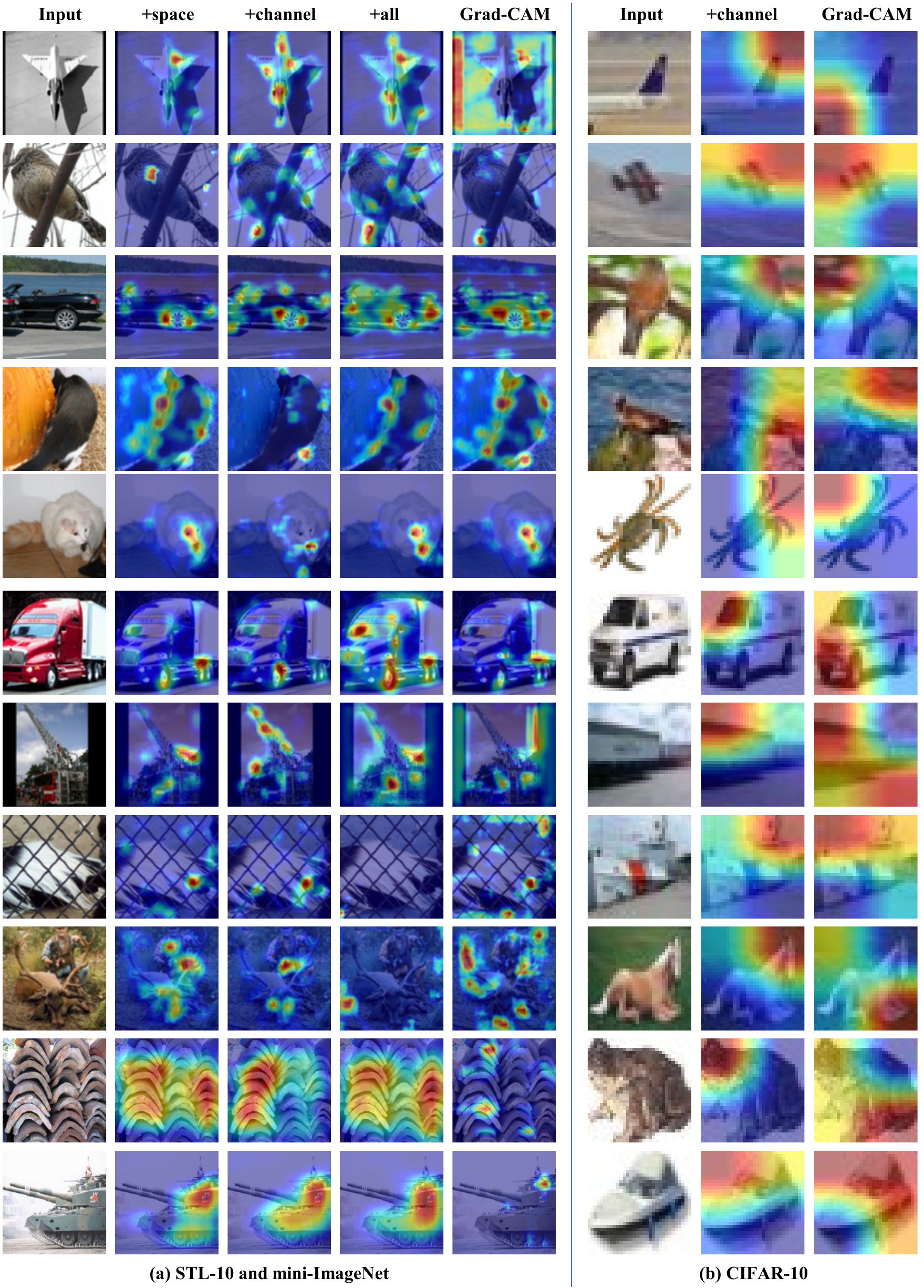}
\caption{More visual results on mini-ImageNet ($224\times 224$), STL-10 ($96\times 96$), and CIFAR-100 ($32\times 32$).  }
\label{visualResultsSupplements}
\end{figure*}

In this section, Fig. \ref{visualResultsSupplements} shows more visual results of GoogLeNet on mini-ImageNet ($224\times 224$), STL-10 ($96\times 96$), and CIFAR-100 ($32\times 32$).
From Fig. \ref{visualResultsSupplements}, we can see that the space-wise constraint successfully restrains the irrelevant background areas for most samples.
What's more, there are fewer wrong responses in the visual results of channel-wise constraint,
which indicates that channel-wise constraint can also restraint irrelevant areas features.
On the contrary, the association maps calculated with Grad-CAM have many inaccurate responses in the background areas,
which means that  Grad-CAM is not accurate for visualizing the classification results, especially for low-confidence samples.
The proposed Model Doctor not only can improve the classification accuracy but also regularize the association between the category and the input features,
which is benefit for observing the classification results.

\end{document}